\documentclass[10pt,twocolumn,letterpaper]{article}

\usepackage{iccv}
\usepackage{times}
\usepackage{epsfig}
\usepackage{graphicx}
\usepackage{amsmath}
\usepackage{amssymb}

\usepackage{multirow}
\usepackage{array}
\usepackage{booktabs}
\usepackage{color}
\usepackage{pifont}
\usepackage{mathtools}
\usepackage{colortbl}
\usepackage{xcolor}
\usepackage{array}
\usepackage{color}
\usepackage{bbding}
\usepackage{makecell}
\usepackage{threeparttable}
\usepackage{float}

\definecolor{myy}{RGB}{126,95,0}
\definecolor{mygray}{gray}{.9}
\definecolor{mygray2}{gray}{0.7}
\definecolor{bblue}{RGB}{30,80,120}
\definecolor{mygray1}{gray}{.8}

\usepackage[breaklinks=true,bookmarks=false]{hyperref}
\iccvfinalcopy 

\def\iccvPaperID{****} 

\begin{document}

\title{Recognize Anything: A Strong Image Tagging Model}

\author{Youcai Zhang$^{*1}$, Xinyu Huang$^{*1}$, Jinyu Ma$^{*1}$, Zhaoyang Li$^{*1}$, Zhaochuan Luo$^{1}$, Yanchun Xie$^{1}$,\\ Yuzhuo Qin$^{1}$, Tong Luo$^{1}$, Yaqian Li$^{1}$, Shilong Liu$^{2}$, Yandong Guo$^{3}$, Lei Zhang$^{2}$\\ 
$^{1}$OPPO Research Institute, $^{2}$International Digital Economy Academy (IDEA), $^{3}$AI$^2$ Robotics\\
$^{*}$Equal Contribution\\
{\tt\small (zhangyoucai,huangxinyu2,majinyu,lichaoyang1)@oppo.com}
\and
}

\maketitle
\ificcvfinal\thispagestyle{empty}\fi



\newlength\savewidth\newcommand\shline{\noalign{\global\savewidth\arrayrulewidth
  \global\arrayrulewidth 1pt}\hline\noalign{\global\arrayrulewidth\savewidth}}
  
\newcommand{\cmark}{\ding{51}}%

\newcommand{\smaller}[1]{\fontsize{7pt}{1em}\selectfont{#1}}

\definecolor{xinyu}{rgb}{0.5,0.8,0.7}
\definecolor{pink}{RGB}{219, 41, 145}

\makeatletter
\newcommand{\algorithmfootnote}[2][\footnotesize]{%
  \let\old@algocf@finish\@algocf@finish
  \def\@algocf@finish{\old@algocf@finish
    \leavevmode\rlap{\begin{minipage}{\linewidth}
    #1#2
    \end{minipage}}%
  }%
}

\renewcommand{\thefootnote}{\fnsymbol{footnote}}

\definecolor{darkergreen}{RGB}{21, 152, 56}
\definecolor{red2}{RGB}{252, 54, 65}
\newcommand\redp[1]{\textcolor{red2}{(#1)}}
\newcommand\greenp[1]{\textcolor{darkergreen}{(#1)}}

\definecolor{pearDark}{HTML}{2980B9}
\definecolor{pearDarker}{HTML}{1D2DEC}

\begin{abstract}
We present the Recognize Anything Model~(RAM): a strong foundation model for image tagging. 
RAM makes a substantial step for large models in computer vision, demonstrating the zero-shot ability to recognize any common category with high accuracy.
RAM introduces a new paradigm for image tagging, leveraging large-scale image-text pairs for training instead of manual annotations. 

The development of RAM comprises four key steps. Firstly, annotation-free image tags are obtained at scale through automatic text semantic parsing. Subsequently, a preliminary model is trained for automatic annotation by unifying the caption and tagging tasks, supervised by the original texts and parsed tags, respectively. Thirdly, a data engine is employed to generate additional annotations and clean incorrect ones. Lastly, the model is retrained with the processed data and fine-tuned using a smaller but higher-quality dataset.

We evaluate the tagging capabilities of RAM on numerous benchmarks and observe impressive zero-shot performance, significantly outperforming CLIP and BLIP. Remarkably, RAM even surpasses the fully supervised manners and exhibits competitive performance with the Google tagging API. We are releasing the RAM at \textcolor{pink}{\url{https://recognize-anything.github.io/}} to foster the advancements of large models in computer vision.




\end{abstract}
\section{Introduction}

\begin{figure}[!htbp]
\begin{center}
\includegraphics[width=0.98\linewidth]{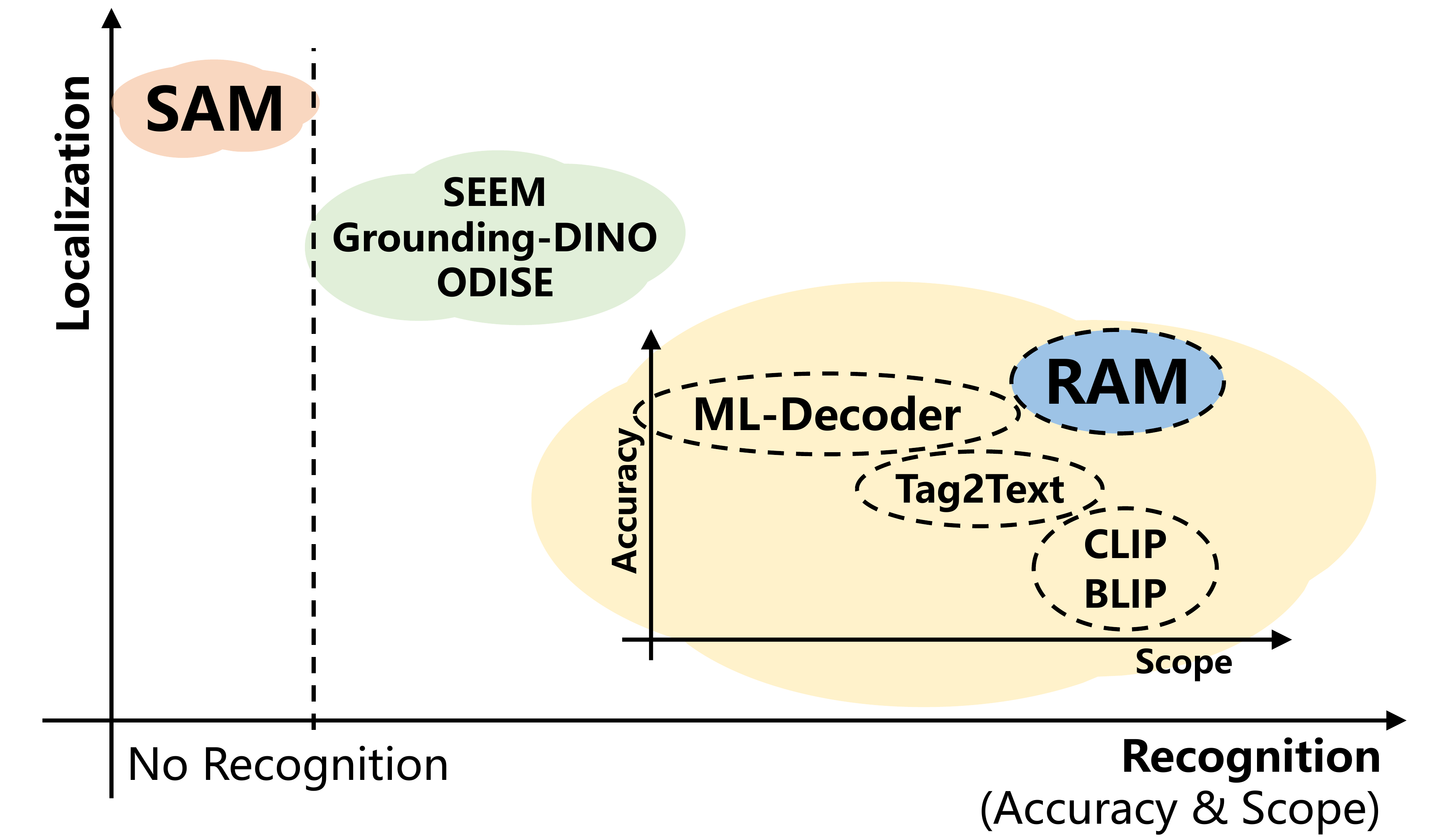}
\end{center}
\caption{SAM excels in providing strong localization capabilities, while it falls short when it comes to recognition tasks. In contrast, RAM exhibits exceptional recognition abilities, surpassing existing models in terms of both accuracy and scope.}
\label{fig:localization_and_recognition}
\end{figure}

\begin{figure*} [!htbp]
\centering
  \includegraphics[width=0.95\linewidth]{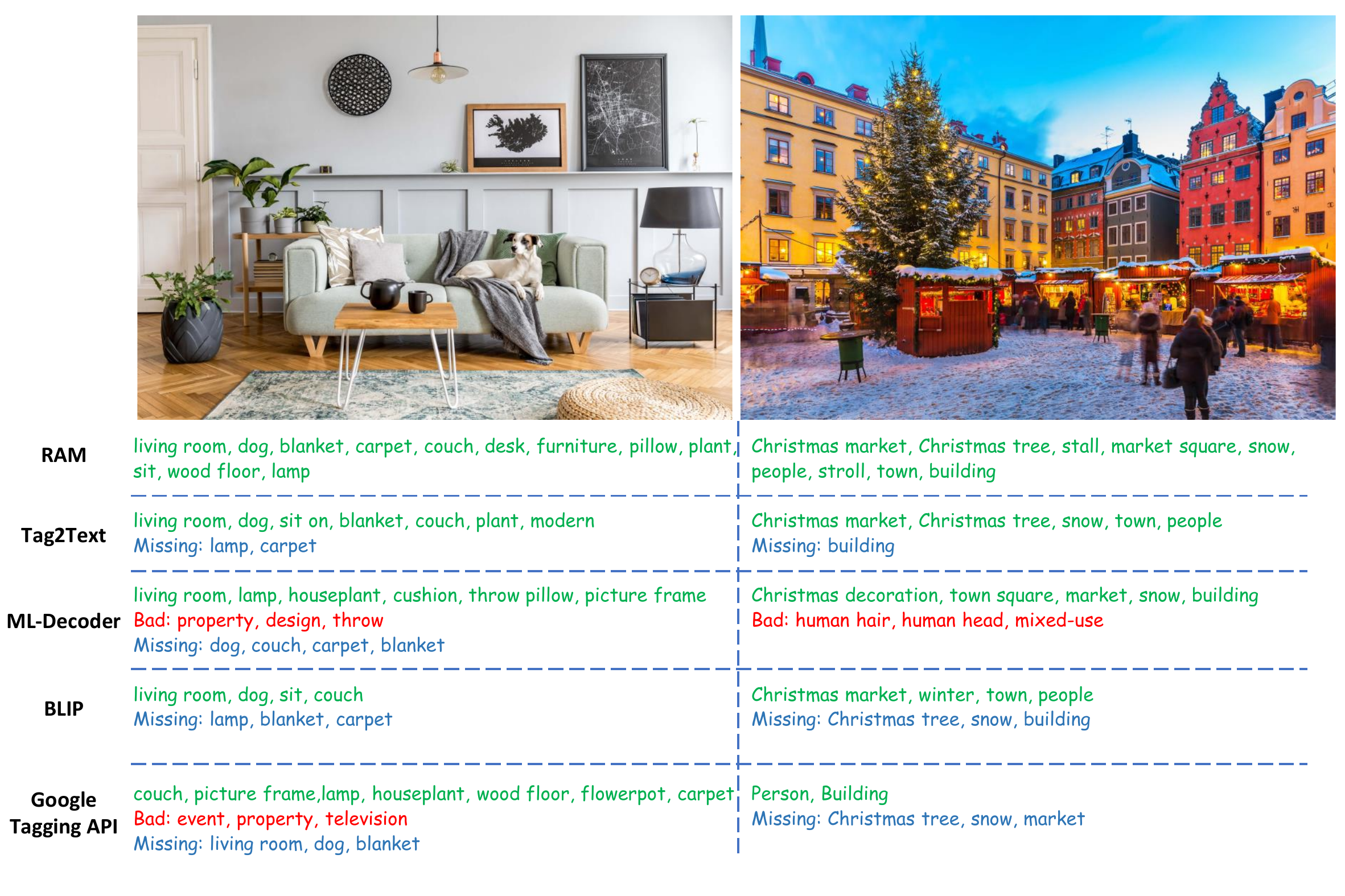}
  \caption{The comparison of recognition capability among tagging models. RAM recognize more valuable tags than other models without missing important part. ML-Decoder and Google tagging API tend to output redundant tags (\textit{e.g.,} \textit{``human head''}) or less relevant tags (\textit{e.g.,} \textit{``property''}) tags. BLIP's tag recall is limited as it relies on caption generation. Note: borderline tags are not listed here.}
  \label{fig:tagging_results}
\end{figure*}

Large language models~(LLM) trained on large-scale web datasets have sparked a revolution in nature language processing~(NLP).
These models\cite{openai2023gpt4,brown2020language} exhibit impressive zero-shot generalization, enabling them to generalize to tasks and data distributions beyond their training domain. 
When it comes to computer vision~(CV), Segment Anything Model~(SAM)~\cite{kirillov2023segany} has also demonstrated remarkable zero-shot localization abilities through data scaling-up.

However, SAM lacks the capability to output semantic labels, which is another foundational task on par with localization.
Multi-label image recognition, also known as image tagging, aims to provide semantic labels by recognizing multiple labels of a given image.
Image tagging is a significant and practical computer vision task, as images inherently contain multiple labels encompassing objects, scenes, attributes, and actions. 
Regrettably, existing models in multi-label classification, detection, segmentation, and vision-language approaches have exhibited deficiency in tagging, characterized by limited scopes or poor accuracy, as illustrated in Figure~\ref{fig:localization_and_recognition}.



Two core components impede the progress of image tagging.
{\it{1)}} The difficulty lies in collecting large-scale high-quality data. Specifically, there is a lack of a universal and unified label system and an efficient data annotation engine, capable of semi-automatic or even automatic annotation of large-scale images with a vast number of categories.
{\it{2)}} There is a lack of efficient and flexible model design that can leverage large-scale weakly-supervised data to construct an open-vocabulary and powerful model. 

To address these key bottlenecks, this paper introduces the Recognize Anything Model~(RAM), a strong foundation model for image tagging.
RAM overcomes the challenges related to data, including label system, dataset and data engine, as well as the limitations in model design.

\textbf{Label System:}
We begin by establishing a universal and unified label system. 
We incorporate categories from popular academic datasets (classification, detection, and segmentation) as well as commercial tagging products (Google, Microsoft, Apple). 
Our label system is obtained by merging all the public tags with the common tags from texts, thus covering most of common labels with a moderate amount of 6,449.
The remaining open-vocabulary labels can be identified through open-set recognition.

\begin{figure*} [t]
\centering
  \includegraphics[width=0.75\linewidth]{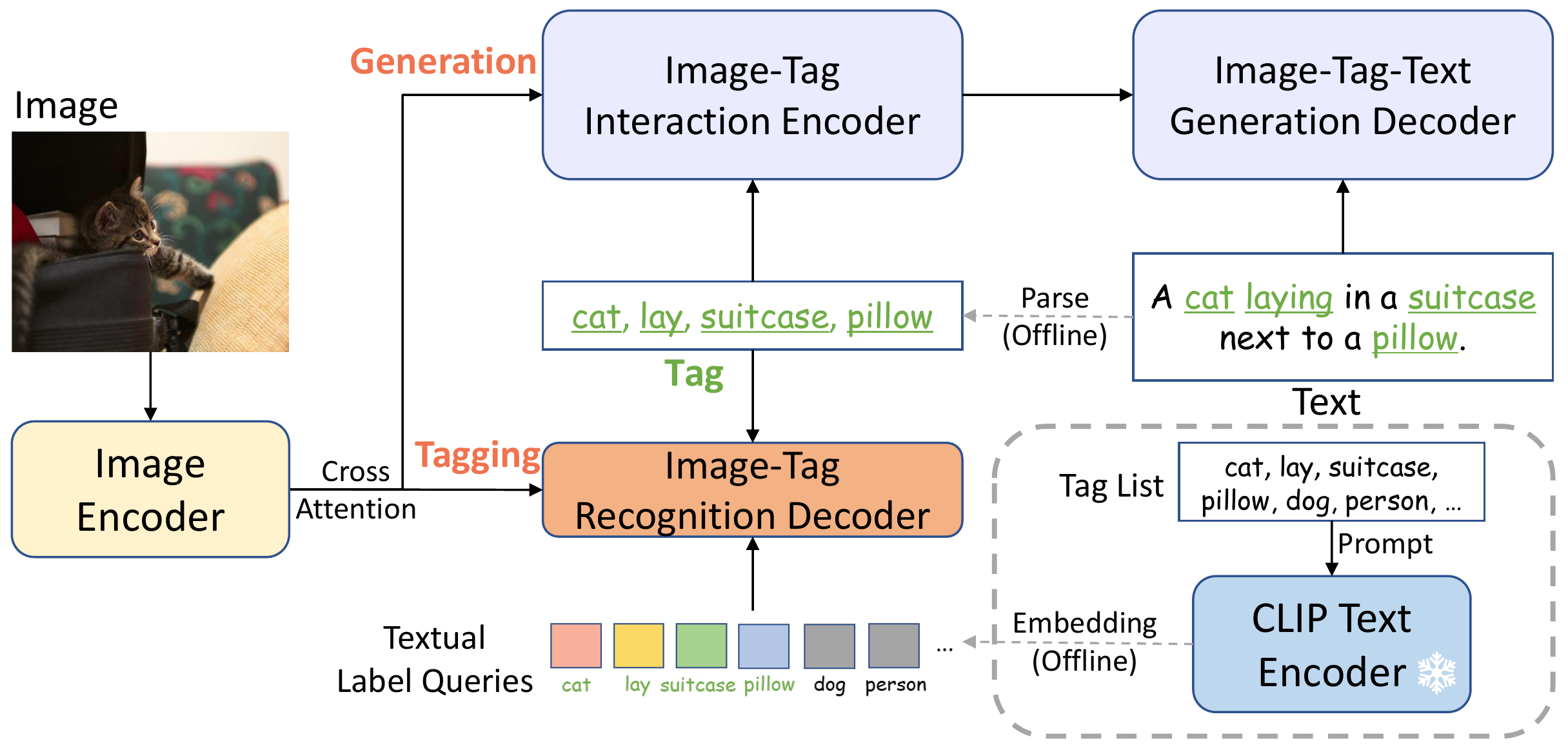}
  \caption{Illustration of RAM's model architecture. Large-scale image tags are obtained from image-text pairs through automatic text semantic parsing. With image-tag-text triplets, RAM unifies the captioning and tagging tasks. Furthermore, RAM introduces an off-the-shelf text encoder to encoder tags into textual label queries with semantically-rich context, empowering the generalization to unseen categories in training stage. }
  \label{fig:model-architercture}
\end{figure*}

\textbf{Dataset:}
How to automatically annotate large-scale images with the label system is another challenge~\cite{zhang2021simple}.
Drawing inspiration from CLIP~\cite{radford2021learning} and ALIGN~\cite{jia2021scaling}, which leverage publicly available image-text pairs at scale to train powerful visual models, we adopt similar datasets for image tagging.
To utilize these large-scale image-text data for tagging, following \cite{huang2022idea,tag2text}, we parse the texts and obtain the image tags through automatic text semantic parsing.
This process allows us to obtain a diverse collection of annotation-free image tags in accordance with image-text pairs. 

\textbf{Data Engine:}
However, the image-text pairs from the web are inherently noisy, often containing missing or incorrect labels.
To enhance the quality of annotations, we design a tagging data engine.
In addressing missing label, we leverage existing models to generate additional tags. 
With regards to incorrect labels, we first localize specific regions corresponding to different tags within the image.
Subsequently, we employ region clustering techniques to identify and eliminate outliers within the same class.
Furthermore, we filter out tags that exhibit contrary predictions between whole images and their corresponding regions, ensuring a cleaner and more accurate annotation.

\textbf{Model:}
Tag2Text~\cite{tag2text} has demonstrated superior image tagging capabilities by the integration of image tagging and caption, employing a lightweight recognition decoder~\cite{liu2021query2label} in conjunction with the original image encoder.
However, the effectiveness of Tag2Text is limited to recognizing fixed and predefined categories.
In contrast, RAM enable generalization to previously unseen categories by incorporating semantic information into label queries.
This model design allows RAM to empower the recognition capabilities of any visual dataset, underlining its potential for diverse applications.


Benefitting from the large-scale, high-quality image-tag-text data and the synergistic integration of tagging with caption, we develop a strong recognize anything model~(RAM).
RAM represents a new paradigm for image tagging, demonstrating that a general model trained on noisy, annotation-free data can outperform fully supervised models. 
The advantages of RAM are summarized as follows:
\begin{itemize}
    \item \textbf{Strong and general.} RAM exhibits exceptional image tagging capabilities with powerful zero-shot generalization as illustrated in Figure~\ref{fig:tagging_results}; 
    \item \textbf{Reproducible and affordable.} RAM requires Low reproduction cost with open-source and annotation-free dataset. Moreover, the strongest version of RAM only requires 3-days 8 A100 GPUs training;
    \item \textbf{Flexible and versatile.}
    RAM offers remarkable flexibility, catering to various application scenarios. By selecting specific classes, RAM can be directly deployed to address specific tagging needs. Furthermore, when combined with localization model~(Grounding DINO and SAM), RAM forms a strong and general pipeline for visual semantic analysis.
\end{itemize}
\begin{figure*} [!htbp]
\centering
  \includegraphics[width=0.55\linewidth]{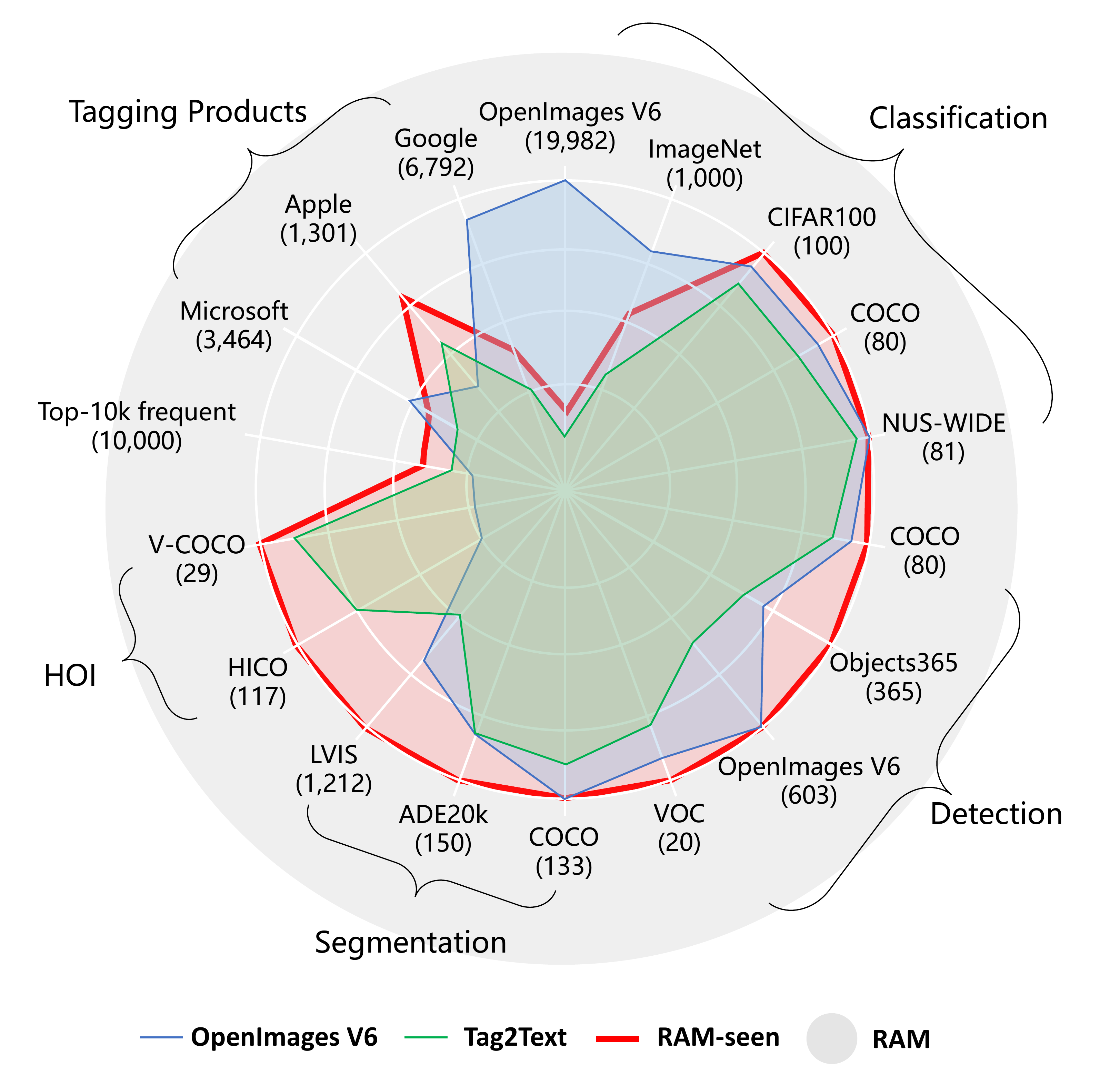}
  \caption{Recognition Scopes of different tagging models. Tag2Text recognizes 3,400+ fixed tags. RAM upgrades the number to 6,400+, covering more valuable categories than OpenImages V6. With open-set capability, RAM is feasible to recognize any common category.}
  \label{fig:category_distribution}
\end{figure*}

\section{Recognize Anything Model}
\subsection{Model Architecture}

As illustrated in Figure~\ref{fig:model-architercture}, we extract image tags through text semantic parsing, providing a large-scale of tags without expensive manual annotations. The overall architecture of RAM is similar to that of Tag2Text\cite{tag2text}, which consists of three key modules: an image encoder for feature extraction, following a image-tag recognition decoder~\cite{liu2021query2label} for tagging, and a text generation encoder-decoder for captioning. The image features interact with tags by the cross-attention layers in the image-tag interaction encoder and recognition decoder. 
In the training stage, the recognition head learns to predict the tags parsed from text, while in the inference stage, it serves as a image-to-tags bridge by predicting tags which provide a more explicit semantic guidance to image captioning. 

Compared with Tag2Text~\cite{tag2text}, RAM's core advancement in model design is the introduction of open-vocabulary recognition. Tag2Text can only recognize the categories that it has seen during training, while RAM can recognize any category.

\subsection{Open-Vocabulary Recognition}

\noindent
\textbf{Textual Label Queries.} 
Inspired by~\cite{ml_decoder, zang2022open}, the pivotal enhancement lies in the incorporation of semantic information into the label queries of the recognition decoder, which facilitates generalization to previously unseen categories in training stage. To achieve this, we utilize an off-the-shelf text encoder to encode the individual tags from the tag list, consequently providing textual label queries with semantically-rich context. In contrast, the label queries employed in the original recognition decode~\cite{tag2text, liu2021query2label} are randomly learnable embeddings, lacking the semantic relationship with unseen categories, thus are confined to predefined seen categories.

\vspace{5pt}
\noindent
\textbf{Implementation Details.} 
We adopt Swin-transformer~\cite{liu2021swin} as the image encoder, as it demonstrated better performance than naive ViT in both vision-language~\cite{tag2text} and tagging domains~\cite{liu2021query2label}. The encoder-decoder used for text generation are 12-layer transformers, and the tag recognition decoder is a 2-layer transformer. 
We utilize the off-the-shelf text encoder from CLIP~\cite{radford2021learning} and perform prompt ensembling~\cite{radford2021learning} to obtain textual label queries. 
We also adopt the CLIP image encoder to distill image feature, which further improves the model's recognition ability for unseen categories via image-text feature alignment.

\subsection{Model Efficiency}

\noindent
\textbf{Training Phase.}
RAM is pretrained on large-scale datasets with a resolution of 224 and fine-tuning at a resolution of 384 using small and high-quality datasets. Empirical evidence suggests that RAM converges rapidly, often with convergence achieved after a minimal number of epochs~(typically less than 5 epochs). This accelerated convergence enhances the reproducibility of RAM with limited computational resources. To illustrate, the version of RAM pretrained on 4 millions necessitate 1-day of computation, and the strongest version of RAM pretrained on 14 million images necessitate a mere 3-days of computation on 8 A100 GPUs.

\vspace{5pt}
\noindent
\textbf{Inference Phase.}
The lightweight image-tag recognition decoder effectively ensures the inference efficiency of RAM on image tagging. Furthermore, we eliminate the self-attention layers from the recognition decoder, which not only further improves efficiency but also circumvents potential interference between label queries. Consequently, instead of fixed categories and quantities, RAM allows customization of label queries for any category and quantity which want to automatically recognize, enhancing its utility across various visual tasks and datasets.

\section{Data}

\subsection{Label System}

This work adopts three guiding principles for the formulation of the label system:
{\it{1)}} Tags that frequently appear in image-text pairs are more valuable due to their representational significance in image description.
{\it{2)}} A variety of domains and contexts should be represented in the tags. Our conception of a \textit{tag} includes objects, scenes, attributes, and actions from a range of sources, which aids model generalization to complex, unseen scenarios.
{\it{3)}} The quantity of tags needs to be moderate. Excessive tag numbers can incur heavy annotation costs.

Initially, we parse 14 million sentences from our pre-training datasets into tags by utilizing a SceneGraphParser~\cite{scene_graph_parser} with minor modifications.
We then hand-pick tags from the top-10k most frequently occurring tags.
Our selection intentionally covers tags from numerous popular datasets for classification, detection, and segmentation, as illustrated in Figure~\ref{fig:category_distribution}.
While most are fully covered, exceptions include ImageNet and OpenImages V6, due to their unusual tag presence.
Additionally, we partially cover tags from leading tagging products, which were obtained via public APIs~\cite{google_api,microsoft_api,apple_api} using open-source images.
Consequently, RAM can recognize up to 6449 fixed tags, which is substantially more than Tag2Text~\cite{tag2text}, and includes a higher proportion of valuable tags.
To reduce redundancy, we collected synonyms via various methodologies including manual checks, referring to WordNet~\cite{wordnet}, translating and merging tags, etc.
Tags within the same synonym group are assigned the same tag ID, resulting in 4585 tag IDs in the label system.

\subsection{Datasets}
Similar to BLIP~\cite{blip} and Tag2Text~\cite{tag2text}, we pre-train our model on widely-used open-source datasets.
4 million~(4M) image and 14 million~(14M) image settings are adopted.
The 4M setting includes two human-annotated datasets, COCO~\cite{coco}~(113K images, 557K captions) and Visual Genome~\cite{vg}~(101K images, 822K captions), along with two large-scale web-based datasets, Conceptual Captions~\cite{cc}~(3M images, 3M captions) and SBU Captions~\cite{sbu}~(849K images, 849K captions).
The 14M setting builds upon the 4M, with the addition of Conceptual 12M~\cite{cc}~(10M images, 10M captions).

\subsection{Data Engine}
Given the predominantly open-source nature of our training datasets, which are largely crawled from the Internet, we encounter a non-negligible amount of missing and incorrect labels.
To mitigate this, we design an automatic data engine to generate additional tags and clean erroneous ones.

\vspace{5pt}
\noindent
\textbf{Generation.}
Our initial step involves training a baseline model using the captions and tags parsed from these captions, similar to the approach used in Tag2Text~\cite{tag2text}. We then leverage this baseline model to supplement both the captions and tags, utilizing its generative and tagging capabilities, respectively. The original captions and tags, in conjunction with the generated captions, corresponding parsed tags, and generated tags, are merged to form a temporary dataset. This step significantly expands the number of tags in the 4M image dataset from 12 million to 39.8 million.

\vspace{5pt}
\noindent
\textbf{Cleaning.}
To address the issue of incorrect tags, we initially employ Grounding-Dino~\cite{dino} to identify and crop regions corresponding to a specific category within all images. 
Subsequently, we cluster the regions from this category based on K-Means++~\cite{arthur2007k} and eliminate the tags associated with the outlier 10\%. 
Simultaneously, we also remove tags without the prediction of this specific category using the baseline model. The motivation is that the precision of tagging models can be improved by predicting regions rather than whole images.




\section{Experiment}

\subsection{Experimental Setting}

\noindent
\textbf{Test Benchmarks.}
We conducted a comprehensive evaluation of the models on various popular benchmark datasets across different computer vision tasks, including classification, detection, and segmentation, as summarized in Table~\ref{table:test_datasets}
For classification, we adopt the OpenImages V6~\cite{openimages}, which contains 9605 categories.
However, due to issues of missing labels and incorrect annotations within the OpenImages dataset, we curated two high-quality subsets: OpenImages-common, comprising 214 well-annotated common categories, and OpenImages-rare, consisting of 200 categories not included in our label system for open-set experiments. 
Additionally, to facilitate better zero-shot evaluation, we employed an internal test set known as OPPO-common, which exhibits high annotation quality.

For detection and segmentation datasets, we selected the widely recognized COCO~\cite{coco} and ADE20k~\cite{ade_1,ade_2} datasets. In these datasets, we focused solely on semantic labels as image-level tagging ground-truth, disregarding bounding boxes and masks. 
It is important to note that ADE20k contains plenty of very small ground-truth annotations and ambiguous categories that deviate from mainstream concepts, \textit{e.g.,} \textit{``buffet''}.
Thus, we created a subset of ADE20k called ADE20k-clean by removing a few small targets and ambiguous categories.

\vspace{5pt}
\noindent
\textbf{Evaluation Metrics.}
To assess the performance of the models, we employed various evaluation metrics. Mean Average Precision~(mAP) was used for reporting results in ablation experiments and comparisons with other classification models. For models where mAP was not available, we utilized Precision/Recall metrics and manually adjusted the threshold of different models to ensure comparability across evaluations.

\begin{table}[!htbp]
\centering
\caption{Details of test benchmarks.}
\vspace{3pt}
\label{table:test_datasets}
\resizebox{\linewidth}{!}{
\begin{threeparttable}
\begin{tabular}{cc|ccccc}
    \hline
    Type                                & Dataset               & \#Category    & \#Image   \\
    \hline
    \multirow{3}{*}{Cls.}               & OPPO-common           & 200           & 44,606     \\
    ~                                   & OpenImages-common~\cite{openimages}     & 214           & 57,224     \\
    ~                                   & OpenImages-rare~\cite{openimages}     & 200           & 21,991     \\
    \hline
    Det.                                & COCO-80~\cite{coco}                  & 80            & 5,000      \\
    \hline
    \multirow{3}{*}{Seg.}               & COCO-133~\cite{coco}                 & 133           & 5,000      \\
    ~                                   & ADE20k~\cite{ade_1,ade_2}                & 150           & 2,000      \\
    ~                                   & ADE20k-clean~\cite{ade_1,ade_2}          & 143           & 2,000      \\
    \hline
\end{tabular}
\end{threeparttable}}
\end{table}

\begin{table*}[!htbp]
    \centering
\caption{Comparison with classification models in mAP. Cells marked with \ding{55} means unable to evaluate on such setting. Cell background color: \colorbox[rgb]{0.93,1.0,0.93}{Green} means fully supervised learning; \colorbox[rgb]{0.92,0.95,1.0}{Blue} means zero-shot performance; \colorbox[rgb]{0.99,0.96,0.9}{Yellow} denotes that the model has seen the corresponding training images, but not the annotations.
Notably, RAM's zero-shot generalization to OpenImages-common is superior to ML-Decoder's full supervision.
RAM can also recognize categories in OpenImages-rare, even though it has not seen them during training. 
}
    \vspace{3pt}
    \label{table:comparation_map}
    \begin{threeparttable}
        \resizebox{0.88\linewidth}{!}{
            \begin{tabular}{l c c c c c c c c c}
                \toprule[1pt]
                \multirow{3.5}{*}{Methods}            & \multirow{3.5}{*}{Tags\tnote{‡}} & \multicolumn{3}{c}{Multi-label Classification} & Detection                           & \multicolumn{3}{c}{Segmentation}                                                                                                                                                                    \\
                \cmidrule(lr){3-5} \cmidrule(lr){6-6} \cmidrule(lr){7-9}
                                                      &                              & OPPO                                           & OpenImages                          & OpenImages                          & \multirow{2}{*}{COCO-80}                   & \multirow{2}{*}{COCO-133}                   & \multirow{2}{*}{ADE20k}             & ADE20k                              \\
                                                      &                              & -common                                        & -common                             & -rare                             &                                         &                                         &                                     & -clean                              \\
                \midrule
                ML-Decoder~\cite{ml_decoder}          &      33.9M        & \cellcolor[HTML]{ECF4FF}{82.4\tnote{†}}       & \cellcolor[HTML]{F0FFF0}{85.8}      & \cellcolor[HTML]{F0FFF0}{\textbf{79.5}}      & \cellcolor[HTML]{ECF4FF}{72.8\tnote{†}} & \cellcolor[HTML]{ECF4FF}{\ding{55}}     & \cellcolor[HTML]{ECF4FF}{\ding{55}} & \cellcolor[HTML]{ECF4FF}{\ding{55}} \\
                MKT~\cite{mkt}                        &       0.6M                       & \cellcolor[HTML]{ECF4FF}{78.2}                 & \cellcolor[HTML]{ECF4FF}{77.8}      & \cellcolor[HTML]{ECF4FF}{63.5}      & \cellcolor[HTML]{ECF4FF}{62.9}          & \cellcolor[HTML]{ECF4FF}{51.0}          & \cellcolor[HTML]{ECF4FF}{37.1}      & \cellcolor[HTML]{ECF4FF}{38.4}      \\
                \midrule
                Tag2Text-4M~\cite{tag2text}  &       11.4M                       & \cellcolor[HTML]{ECF4FF}{83.0}                 & \cellcolor[HTML]{ECF4FF}{82.9}      & \cellcolor[HTML]{ECF4FF}{\ding{55}} & \cellcolor[HTML]{FDF5E6}{78.3\tnote{†}} & \cellcolor[HTML]{FDF5E6}{66.9\tnote{†}} & \cellcolor[HTML]{ECF4FF}{\ding{55}} & \cellcolor[HTML]{ECF4FF}{\ding{55}} \\
                Tag2Text-14M~\cite{tag2text} &         33.6M                     & \cellcolor[HTML]{ECF4FF}{85.4}                 & \cellcolor[HTML]{ECF4FF}{83.4}      & \cellcolor[HTML]{ECF4FF}{\ding{55}} & \cellcolor[HTML]{FDF5E6}{78.2\tnote{†}} & \cellcolor[HTML]{FDF5E6}{67.1\tnote{†}} & \cellcolor[HTML]{ECF4FF}{\ding{55}} & \cellcolor[HTML]{ECF4FF}{\ding{55}} \\
                \midrule
                RAM-4M                                &           39.3M                   & \cellcolor[HTML]{ECF4FF}{85.6}                 & \cellcolor[HTML]{ECF4FF}{86.0}      & \cellcolor[HTML]{ECF4FF}{66.7}      & \cellcolor[HTML]{FDF5E6}{79.0}          & \cellcolor[HTML]{FDF5E6}{68.3}          & \cellcolor[HTML]{ECF4FF}{51.5}      & \cellcolor[HTML]{ECF4FF}{53.2}      \\
                RAM-14M                               &        119.9M                      & \cellcolor[HTML]{ECF4FF}{\textbf{86.9}}                 & \cellcolor[HTML]{ECF4FF}{\textbf{86.5}}      & \cellcolor[HTML]{ECF4FF}{69.2}      & \cellcolor[HTML]{FDF5E6}{\textbf{80.6}}          & \cellcolor[HTML]{FDF5E6}{\textbf{69.4}}          & \cellcolor[HTML]{ECF4FF}{\textbf{55.4}}      & \cellcolor[HTML]{ECF4FF}{\textbf{56.9}}      \\
                \bottomrule[0.5pt]
            \end{tabular}
        }
        \begin{tablenotes}
            \footnotesize
            \item[†] A few categories that are not supported by the model are excluded when calculating mAP.
            \item[‡] The total number of common tags that co-occur in the training set and the top-10k parsed tags.
        \end{tablenotes}
    \end{threeparttable}
\end{table*}

\begin{table*}[!htbp]
    \centering
    \caption{Comparison with detection, segmentation and vision-language models in Precision/Recall. Cells marked with \cellcolor[HTML]{ECF4FF}{\ding{81}} means poor performance in large-sized categories, or long inference time due to the high image resolution, \textit{e.g.,} 1024 for ODISE. Notably, RAM outperforms CLIP and BLIP with large margins in common categories.}
    \vspace{3pt}
    \label{table:comparation_pr}
    \begin{threeparttable}
        \resizebox{\linewidth}{!}{
            \begin{tabular}{l c c c c c c c c c}
                \toprule[1pt]
                \multirow{3.5}{*}{Methods}             & \multirow{3.5}{*}{Backbone} & \multicolumn{3}{c}{Multi-label Classification} & Detection                             & \multicolumn{3}{c}{Segmentation}                                                                                                                                                                                        \\
                \cmidrule(lr){3-5} \cmidrule(lr){6-6} \cmidrule(lr){7-9}
                                                       &                              & OPPO                                           & OpenImages                            & OpenImages                            & \multirow{2}{*}{COCO-80}                          & \multirow{2}{*}{COCO-133}                          & \multirow{2}{*}{ADE20k}               & ADE20k                                \\
                                                       &                              & -common                                        & -common                               & -rare                               &                                                &                                                &                                       & -clean                                \\
                \midrule
                Grounding-DINO~\cite{liu2023grounding} & Swin-B                       & \cellcolor[HTML]{ECF4FF}{\ding{81}}            & \cellcolor[HTML]{ECF4FF}{\ding{81}}   & \cellcolor[HTML]{ECF4FF}{\ding{81}}   & \cellcolor[HTML]{F0FFF0}{\textbf{83.1 / 86.9}}          & \cellcolor[HTML]{F0FFF0}{66.4 / 48.3}          & \cellcolor[HTML]{ECF4FF}{34.3 / 24.7} & \cellcolor[HTML]{ECF4FF}{35.6 / 26.0} \\
                \midrule
                ODISE~\cite{odise}                     & Diffusion-v3             & \cellcolor[HTML]{ECF4FF}{\ding{81}}            & \cellcolor[HTML]{ECF4FF}{\ding{81}}   & \cellcolor[HTML]{ECF4FF}{\ding{81}}   & \cellcolor[HTML]{F0FFF0}{78.5 / 85.9}          & \cellcolor[HTML]{F0FFF0}{\textbf{71.1 / 80.2}}          & \cellcolor[HTML]{ECF4FF}{47.4 / 48.0} & \cellcolor[HTML]{ECF4FF}{48.2 / 50.3} \\
                SEEM~\cite{seem}                       & FocalNet-L               & \cellcolor[HTML]{ECF4FF}{\ding{55}}            & \cellcolor[HTML]{ECF4FF}{\ding{55}}   & \cellcolor[HTML]{ECF4FF}{\ding{55}}   & \cellcolor[HTML]{F0FFF0}{75.7 / 67.8}          & \cellcolor[HTML]{F0FFF0}{71.8 / 61.0}          & \cellcolor[HTML]{ECF4FF}{\ding{55}}   & \cellcolor[HTML]{ECF4FF}{\ding{55}}   \\
                \midrule
                CLIP-400M~\cite{radford2021learning}        & ViT-B                     & \cellcolor[HTML]{ECF4FF}{76.6 / 54.1}          & \cellcolor[HTML]{ECF4FF}{77.9 / 52.9} & \cellcolor[HTML]{ECF4FF}{\textbf{67.5 / 46.5}} & \cellcolor[HTML]{ECF4FF}{64.0 / 38.7}          & \cellcolor[HTML]{ECF4FF}{47.8 / 36.4}          & \cellcolor[HTML]{ECF4FF}{30.3 / 5.3}  & \cellcolor[HTML]{ECF4FF}{31.0 / 5.5}  \\
                BLIP-129M~\cite{blip}                       & ViT-B                             & \cellcolor[HTML]{ECF4FF}{76.7 / 57.5}          & \cellcolor[HTML]{ECF4FF}{78.6 / 55.1} & \cellcolor[HTML]{ECF4FF}{65.2 / 46.5} & \cellcolor[HTML]{FDF5E6}{67.0 / 39.0}          & \cellcolor[HTML]{FDF5E6}{53.8 / 34.6}          & \cellcolor[HTML]{ECF4FF}{28.5 / 8.8}  & \cellcolor[HTML]{ECF4FF}{29.1 / 9.3}  \\ \hline
                Tag2Text-4M~\cite{tag2text}   & Swin-B                       & \cellcolor[HTML]{ECF4FF}{76.6 / 74.8}          & \cellcolor[HTML]{ECF4FF}{75.9 / 71.9} & \cellcolor[HTML]{ECF4FF}{\ding{55}}   & \cellcolor[HTML]{FDF5E6}{80.5 / 66.1\tnote{†}} & \cellcolor[HTML]{FDF5E6}{71.2 / 54.0\tnote{†}} & \cellcolor[HTML]{ECF4FF}{\ding{55}}   & \cellcolor[HTML]{ECF4FF}{\ding{55}}   \\
                Tag2Text-14M~\cite{tag2text}  & Swin-B                       & \cellcolor[HTML]{ECF4FF}{77.9 / 79.4}          & \cellcolor[HTML]{ECF4FF}{76.4 / 73.3} & \cellcolor[HTML]{ECF4FF}{\ding{55}}   & \cellcolor[HTML]{FDF5E6}{80.1 / 64.5\tnote{†}} & \cellcolor[HTML]{FDF5E6}{71.2 / 53.2\tnote{†}} & \cellcolor[HTML]{ECF4FF}{\ding{55}}   & \cellcolor[HTML]{ECF4FF}{\ding{55}}   \\
                \midrule
                RAM-4M                                 & Swin-B                       & \cellcolor[HTML]{ECF4FF}{78.4 / 75.2}          & \cellcolor[HTML]{ECF4FF}{79.2 / 73.7} & \cellcolor[HTML]{ECF4FF}{53.9 / 48.4} & \cellcolor[HTML]{FDF5E6}{81.8 / 66.1}          & \cellcolor[HTML]{FDF5E6}{74.3 / 54.0}          & \cellcolor[HTML]{ECF4FF}{47.0 / 47.6} & \cellcolor[HTML]{ECF4FF}{47.8 / 50.3} \\
                RAM-14M                                & Swin-L                       & \cellcolor[HTML]{ECF4FF}{\textbf{78.8 / 79.4}}          & \cellcolor[HTML]{ECF4FF}{\textbf{80.3 / 75.7}} & \cellcolor[HTML]{ECF4FF}{53.8 / 54.3} & \cellcolor[HTML]{FDF5E6}{82.9 / 66.4}          & \cellcolor[HTML]{FDF5E6}{74.3 / 54.1}          & \cellcolor[HTML]{ECF4FF}{\textbf{53.2 / 50.0}} & \cellcolor[HTML]{ECF4FF}{\textbf{53.7 / 52.2}} \\
                \bottomrule[0.5pt]
            \end{tabular}
        }
        \begin{tablenotes}
            \footnotesize
            \item[†] A few categories that are not supported by the model are excluded when calculating precision and recall.
        \end{tablenotes}
    \end{threeparttable}
\end{table*}

\begin{table*}[]
\begin{center}
\centering
\caption{Ablation study of RAM model based on Tag2Text baselines. \textit{``Seen Categories''} refers to the number of training categories.\textit{``Captioning''} refers to the joint training of captioning and tagging tasks. \textit{``Textual Queries''} refers to using a text encoder to generate label queries possessing semantic information. \textit{``Distillation''} refers to image feature distillation using CLIP's image encoder. }
\vspace{3pt}
\label{table:model_ablation}
\begin{tabular}{c|c|ccc|ccc}
\hline
\multirow{2}{*}{Case}     & \multirow{2}{*}{\begin{tabular}[c]{@{}c@{}}Seen \\ Categories\end{tabular}} & \multirow{2}{*}{Captioning} & \multirow{2}{*}{\begin{tabular}[c]{@{}c@{}}Textual \\ Queries\end{tabular}} & \multirow{2}{*}{Distillation} &OPPO & \multicolumn{2}{c}{OpenImages} \\
&    &    &          &   & -common & -common & -rare \\ \hline
\multirow{2}{*}{Tag2Text} & 3,429  & &     &   & 80.60  & 83.52 & \ding{55}   \\
                          & 3,429                                                                        & \ding{51}  &                     &   &  81.37  & 84.04  & \ding{55}                       \\ \hline
(a)                       & 3,429   & \ding{51}  & \ding{51}   &   &  81.22   & 84.09     & 60.99  \\
(b)                       & 3,429   & \ding{51}  & \ding{51}   & \ding{51}   &  81.70 & 84.16       & 61.88                   \\
(c)                       & \textbf{6,449}   & \ding{51}  & \ding{51}  & \ding{51}     & 80.27  & 83.09       & 63.54 \\ \hline
\end{tabular}
\end{center}
\end{table*}


\begin{table*}[]
\begin{center}
\caption{Ablation study of data engine. \textit{``Parsing''} means the training tags parsed from the captions. \textit{``Generation''} means the supplementation of captions and tags. \textit{``Cleaning''} refers to data cleaning. \textit{``Fine-tuning''} refers to fine-tuning the pre-trained model with COCO.}
\vspace{3pt}
\label{table:data_ablation}
\begin{threeparttable}
\begin{tabular}{l|cc|cccc|ccc}
\hline

\multirow{2}{*}{Backbone} & \multicolumn{2}{c|}{Pre-train}  & \multirow{2}{*}{Parsing} & \multirow{2}{*}{Generation}  & \multirow{2}{*}{Cleaning} & \multirow{2}{*}{Fine-tuning} &OPPO & \multicolumn{2}{c}{OpenImages} \\

& \#Images   &  \#Tags  &      &    &  & & -common & -common & -rare \\ \hline

\multirow{5}{*}{Swin-Base}  & 4M & 12.0M & \ding{51}       &            &            &             & 80.27    & 83.09    &  63.54  \\
                       & 4M  & 41.7M   & \ding{51}       & \ding{51}  &            &             & 82.50    & 84.27    &  67.17   \\
                       & 4M  & 39.8M    & \ding{51}       & \ding{51}  &  \ding{51} &             & 82.83    & 84.94    &  66.88   \\
                       & 4M  & 39.8M & \ding{51}       & \ding{51}  &   \ding{51} & \ding{51}  & 85.56    &  86.01   &  66.74    \\
                       & 14M  &  121.5M & \ding{51}          & \ding{51}       &     \ding{51}    &     &    83.52   &   85.39     & 68.54  \\ 
                         & 14M &  121.5M& \ding{51}          & \ding{51}        &        \ding{51}    &   \ding{51}                                                     &      86.47                                                        & 86.50                                            &  68.79                \\

                       \hline
\multirow{2}{*}{Swin-Large} & 14M        &    121.5M    & \ding{51}          & \ding{51}        &      \ding{51}      &                                                        &         83.26                                                     & 84.94 &   68.60                                                          \\
                       &     14M   & 121.5M  & \ding{51}          & \ding{51}        & \ding{51}          &     \ding{51}                                                   &                                                    86.92          &   86.46& 69.21                                                          \\ \hline
\end{tabular}

\end{threeparttable}
\end{center}
\end{table*}

\subsection{Comparison with SOTA Models}
\noindent
\textbf{Comparison with Multi-Label Classification Models.}
We compare RAM with state-of-the-art~(SOTA) models in multi-label classification, as show in Table~\ref{table:comparation_map}.
Generally, a generalist model typically lacks expertise in specific domains, whereas an expert model struggles to generalize beyond its specialized field. 
Specifically, the supervised expert model ML-Decoder~\cite{ml_decoder} excels in its designated domain of expertise, OpenImages, but faces challenges in generalizing to other domains and unseen categories.
MKT~\cite{mkt} is a generalist model in tagging by transferring the knowledge from CLIP, fails to achieve satisfactory accuracy across all domains.
Tag2Text~\cite{tag2text} is powerful at zero-shot tagging, but it lacks the ability to handle open-set scenarios.

RAM exhibits impressive tagging abilities, showcasing impressive accuracy and broad coverage. 
Particularly noteworthy is the performance of RAM-4M, which surpasses ML-Decoder on the OpenImages-common dataset. While ML-Decoder relies on 9 million annotated images from OpenImages, our RAM-4M achieves higher accuracy with a training set of 4 million annotation-free image-text data. This improvement is attributed to the utilization of 39.3 million common tags derived from the 4 million images, outperforming ML-Decoder trained with only 33.9 million common tags from 9 million images.
Moreover, RAM can recognize any common category by leveraging a vast range of 6,400+ seen common categories, coupled with its open-vocabulary ability.

\vspace{5pt}
\noindent
\textbf{Comparison with Detection and Segmentation Models.}
The comparison in Table \ref{table:comparation_pr} reveals that supervised detection and segmentation models excel in specific domains such as COCO datasets, which encompass a limited number of categories. However, these models face challenges when it comes to recognizing a larger number of categories.
On the one hand, they take much more computational overheads as they requires more complex network and larger input image sizes for extra localization task. 
Especially, ODISE~\cite{odise} takes long inference time due to its adoption of the diffusion model and large input image resolution.
On the other hand, the scalability of training data for detection and segmentation is limited, resulting in poor generalization performance for these models. 
Although Grounding-DINO~\cite{liu2023grounding} serve as a generalist model, it struggles to achieve satisfactory performance for large-sized categories.
In contrast, RAM demonstrates impressive open-set ability, surpassing existing detection and segmentation models. RAM showcases its capability to generalize across a broader range of categories, providing a robust solution for the challenges faced by conventional detection and segmentation models.

\vspace{5pt}
\noindent
\textbf{Compared with Vision-Language Models.}
Despite the open-set recognition capabilities of CLIP~\cite{radford2021learning} and BLIP~\cite{blip}, these models suffer from subpar accuracy. Furthermore, their interpretability is limited, as they rely on cosine similarity computations of dense embeddings for image-text pairs.
In contrast, RAM exhibits superior performance, surpassing CLIP and BLIP by a significant margin, with accuracy increases of over 20\% observed across almost all datasets. However, it is worth noting that RAM performs slightly worse than CLIP and BLIP in the case of OpenImages-rare dataset. We attribute this discrepancy to the smaller training dataset utilized for RAM and the relatively less emphasis placed on rare classes during training.

\subsection{Model Ablation Study}
In Table~\ref{table:model_ablation}, we study the impact of various model improvements to RAM based on Tag2Text~\cite{tag2text} and make the following key observations.
{\it{1)}} The training integration of captioning and tagging can promote the tagging ability. {\it{2)}} The open-set recognition capability can be achieved through textual queries by CLIP~\cite{radford2021learning}, but has little impact on the seen categories in training. {\it{3)}} The expansion of the label system introduces minimal impact on existing categories, which can be attributed to the additional categories increases the difficulty of model training. However, this expansion concurrently enhances the model's coverage and enhances the open-set ability of unseen categories. 

\subsection{Data Engine Ablation Study}

We present an ablation study of the data engine in Table \ref{table:data_ablation}.
The findings are summarized as follows:
{\it{1)}} 
Adding more tags from 12.0M to 41.7M significantly improves model performance across all test sets, indicating the severe missing label problem in original datasets.
{\it{2)}} 
Further cleaning the tags of some categories results in a slight increase in performance on the OPPO-common and OpenImages-common test sets.
Limited by the inference speed of Grounding-Dino, we only conduct cleaning process for 534 categories.
{\it{3)}} Scaling up the training images from 4M to 14M brings remarkable improvements across all test sets.
{\it{4)}} Employing a larger backbone network leads to a slight improvement on OpenImages-rare and even slightly inferior performance on common categories. We attribute this phenomenon to our insufficient resources available for conducting hyper-parameter search.
{\it{5)}} Fine-tuning with tags parsed from the COCO Caption dataset~\cite{coco} demonstrates remarkable increases in performance on the OPPO-common and OpenImages-common test sets. 
The COCO Caption dataset provides five descriptive sentences for each image, offering a comprehensive description that approximates a complete set of tag labels.

\section{Conclusion}
We present the Recognize Anything Model (RAM), a strong foundation model designed for image tagging, which heralds a novel paradigm in this field. 
RAM demonstrate the zero-shot ability to recognize any category with high accuracy, surpassing the performance of both fully supervised models and existing generalist approaches like CLIP and BLIP. RAM represents a considerable advancement for large-scale models in the field of computer vision, holding the potential to empower the recognition capabilities of any visual tasks or datasets.  

There still exists room for further refinement of RAM. For example, scaling up the training dataset beyond 14 million images to better cover diverse domains, multiple rounds of data engine, increasing the backbone parameters to enhance the model capacity.




\vspace{5pt}
\noindent
\textbf{Limitations.}
Similar to CLIP, the current version of RAM efficiently recognizes common objects and scenes, yet struggles with abstract tasks like object counting. Moreover, zero-shot RAM's performance lags behind task-specific models in fine-grained classifications, such as differentiating between car models or identifying specific flower or bird species. It is also noteworthy that RAM is trained on open-source datasets and could potentially reflect dataset biases.



{\small
\bibliographystyle{ieee_fullname}
\bibliography{egbib}
}

\end{document}